\documentclass[journal]{IEEEtran}

\usepackage{xcolor,soul,framed} 

\colorlet{shadecolor}{yellow}
\usepackage[pdftex]{graphicx}
\graphicspath{{../pdf/}{../jpeg/}}
\DeclareGraphicsExtensions{.pdf,.jpeg,.png}

\usepackage[cmex10]{amsmath}
\usepackage{array}
\usepackage{mdwmath}
\usepackage{mdwtab}
\usepackage{eqparbox}
\usepackage{url}

\begin{document}

\title{I-SAFE: Instant Suspicious Activity identiFication at the Edge using Fuzzy Decision Making}

\author{
\IEEEauthorblockN{Seyed Yahya Nikouei${^a}$, Yu Chen${^a}$, Alexander Aved$^{b}$, Erik Blasch$^{b}$, Timothy R. Faughnan${^c}$}

\IEEEauthorblockA{${^a}$Dept. of Electrical \& Computer Engineering, Binghamton University, SUNY,  Binghamton, NY 13902, USA \\ $^{b}$The U.S. Air Force Research Laboratory, Rome, NY 13441, USA\\
${^c}$New York State University Police, Binghamton University, SUNY, Binghamton, NY 13902, USA\\
\{snikoue1, ychen, tfaughn\}@binghamton.edu, \{alexander.aved, erik.blasch\}@us.af.mil}
}

\maketitle

\begin{abstract}
Urban imagery usually serves as forensic analysis and by design is available for incident mitigation. As more imagery collected, it is harder to narrow down to certain frames among thousands of video clips to a specific incident. A real-time, proactive surveillance system is desirable, which could instantly detect dubious personnel, identify suspicious activities, or raise momentous alerts. The recent proliferation of the edge computing paradigm allows more data-intensive tasks to be accomplished by smart edge devices with lightweight but powerful algorithms. This paper presents a forensic surveillance strategy by introducing an Instant Suspicious Activity identiFication at the Edge (I-SAFE) using fuzzy decision making. A fuzzy control system is proposed to mimic the decision-making process of a security officer. Decisions are made based on video features extracted by a lightweight Deep Machine Learning (DML) model. Based on the requirements from the first-line law enforcement officers, several features are selected and fuzzified to cope with the state of uncertainty that exists in the officers’ decision-making process. Using features in the edge hierarchy minimizes the communication delay such that instant alerting is achieved. Additionally, leveraging the Microservices architecture, the I-SAFE scheme possesses good scalability given the increasing complexities at the network edge. Implemented as an edge-based application and tested using exemplary and various labeled dataset surveillance videos, the I-SAFE scheme raises alerts by identifying the suspicious activity in an average of 0.002 seconds. Compared to four other state-of-the-art methods over two other data sets, the experimental study verified the superiority of the I-SAFE decentralized method.
\end{abstract}

\begin{IEEEkeywords}
Edge Computing, Smart Surveillance, Fuzzy Control System, Object Detection, Behaviour Identification.
\end{IEEEkeywords}

\IEEEpeerreviewmaketitle

\section{Introduction}
\label{intro}

Providing safety and well-being of the residents who live in populated cities is a rising challenge. City planners answered the security challenge by adding more cameras to enhance ubiquitous surveillance \cite{chen2017enabling}. The capability of real-time activities monitoring enables faster reaction for first responders in cases of emergencies. For instance, when a security agent sees the live footage and identifies a problem, quick actions can be taken. However, it is very difficult, if not impossible, for a security agent to focus on one of so many cameras when an event happens. In fact, most surveillance video streams are normally used as a forensics  for diagnostics, lessons learned, and preparation for future events. Likewise, it takes long time to investigate information from thousands of video clips. The situation could be worse when the footage has been deleted due to limited storage space. To utilize the limited storage space more efficiently, many video surveillance systems integrate motion sensors. The cameras do not capture videos unless they are triggered by the motion sensors’ detection of certain movements. 

Currently, smart security cameras use intelligence to detect, classify and recognize objects of interest to determine which video clips to retain \cite{ma2017survey}. These advances accelerate the decision making for the operator and also classify the data for later forensic analysis. Recently, machine learning (ML) models are adopted to detect anomalous behaviours by identifying certain bio-mechanical movements \cite{wang2013intelligent}, but suffer from high false positive rates from inadequate training.

Edge computing is recognized as a promising solution to tackle the challenges in today's ubiquitously deployed video surveillance systems \cite{chen2016smart}, \cite{shi2016edge}. Migrating computing power to the edge allows more intelligence at each edge node such that on-site or near-site data processing becomes feasible, which consequently enables real-time object detection \cite{nikouei2018lcnn}, tracking \cite{nikouei2018kerman}, and feature abstraction at the edge. Presently, it is still challenging to recognize the activities based on the features and identify suspicious behaviors. 

Recently, researchers introduced a lightweight Convolutional Neural Network (L-CNN) \cite{nikouei2018lcnn} for real-time object detection focusing on a primary object of interest (humans) and a hybrid KERnelized Machine learning and Artificial Networks (Kerman) algorithm \cite{nikouei2018kerman} for object tracking at the edge. The Kerman algorithm analyzes each frame of the video stream and extracts movement features and models patterns.  

In this paper, we introduce an Instant Suspicious Activity identiFication at the Edge (I-SAFE) using a fuzzy decision making engine. A set of contextualized detectors are created by considering the features in a spatio-temporal context \cite{snidaro2016context}, which include the location of the camera and time of the day. Then, a fuzzy model with five membership functions is proposed to decide whether or not the behavior or activity of each of the people in the frame are suspicious. The fuzzy model was generated with inputs from campus police officers. Their knowledge and experience are integrated in the rule set, which uses the fuzzified contextualized features. The experimental study using real-world surveillance video streams verifies the I-SAFE scheme nominates suspicious activities in average of 0.002 seconds. 

The contributions of this paper are as follows: 

\begin{itemize}
    \item[1)] A lightweight smart safety system which is decentralized and real-time which detects humans and rises an individual based alarm in case certain factors are seen in the behavior;
    
    \item[2)] A data experiment leveraging domain expert features selection that best describe the behavior elements of each pedestrian to support edge paradigm constraints;
    
     \item[3)] A noise resistant fuzzy control module that upon receiving image features determines the intentions of each pedestrian in the frame; and
    
    \item[4)] A comprehensive experimental study that verified the effectiveness of the I-SAFE decentralized method by comparing it to four other state-of-the-art methods over two other data sets.
\end{itemize}

The rest of this paper is structured as follows. In Section \ref{background}, the background knowledge of the problem and fuzzy systems are introduced. Section \ref{sys} presents a novel smart surveillance architecture to support the proposed I-SAFE scheme. Section \ref{data} explains the feature extraction method and contextualization. The fuzzy model of the I-SAFE scheme is introduced in Section \ref{fuzzy}. The experimental results are presented in Section \ref{experimental}. Finally, Section \ref{conc} provides conclusions.

\section{Background and Related Work}
\label{background}

\subsection{Human-in-Loop Surveillance Systems}

The surveillance community is aware of the growing demand for human resources to interpret data such as live video streams \cite{blasch2012high}. The ubiquitous deployment of networked static and mobile cameras creates a huge amount of video data that is being transmitted to data centers for analysis \cite{blasch2016cloud} and atomize the process \cite{ma2017survey}. Many automated object detection algorithms have been investigated using ML~\cite{ribeiro2017study} and statistical analysis~\cite{fuse2017statistical} approaches that are implemented at the server side of the surveillance system. 

There are also efforts made to promote operators’ awareness by leveraging context \cite{blasch2014context}, providing query languages \cite{aved2015multi}, re-configuring the networked cameras \cite{piciarelli2016dynamic}, utilizing event-driven visualization  \cite{fan2017heterogeneous}, and mapping conventional real-time images to 3D camera images \cite{jiewu1}. Lack of scalability is still a challenge in traditional human-in-the-loop solutions to meet the demand of real-time surveillance. 

\subsection{Safety Modeling and Anomaly Detection}

Anomaly detection can have various definitions. For example, some researchers define anomaly as the most rare state in a sequence \cite{del2016discriminative}, \cite{liu2018future}. In case of a video, the algorithm selects the most rare frames in the sequence. For example, detecting a person or vehicle not in a normal behavior for pattern of life/anomaly detection (POL/AD) \cite{blasch2012pattern}. In the Appearance and Motion DeepNet (AMDN) framework, several classifiers work in parallel to detect whether or not there is an object in a frame \cite{xu2015learning}, such as a car or bike in the park \cite{mahadevan2010anomaly}. An integrated pipeline incorporates the output of object trajectory analysis and pixel-based analysis for abnormal behavior inference \cite{cocsar2017toward}. It aims at better automation in the surveillance system with an algorithm that robustly captures activities such as: loitering, fighting and passing cars. Although POL/AD is comprehensive, the approach is unfortunately too expensive to be implemented on fog nodes which host edge units' data streams. Labeling partial video segments rather than bounding boxes in video frames, anomaly analysis segments the video where an action such as moving, stealing, or incident \cite{holloway2014activity}, \cite{sultani2018real}. Video range labeling using a refined Recurrent Neural Network (RNN) also translates to a more accurate rare instance detection along with outputs of the bounding box around the anomalous object  \cite{luo2017revisit}. 

Loitering detection is selected as the case study to implement and test the I-SAFE framework. Loitering is moving back and forth around a centralized spot. Hence stopping, starting, and returning to the same spot are obvious indicators. Although loitering and move-top-move actions look similar, having a model that differentiates between the motivations is helpful to distinguish them accurately. A spatio-temporal clustering method based on the pedestrian speed is adopted for classification \cite{palma2008clustering}. It uses some features for motions detection and alarm raising, which suffers an overall low performance. An unsupervised dynamic sparse coding approach method was suggested for unusual event detection using an atomically learned event dictionary \cite{zhao2011online}. This approach shows the unusual scene which loitering may be one, but does not concern the security aspect. Finally, the loitering problem was tried using a Markov random field (MRF) \cite{kim2009observe} and generally seeks rare occurrence anomaly detection. 

Two-dimensional human pose estimation in still images and videos has been explored \cite{ioffe1999finding}, where both top-down \cite{mikolajczyk2004human} and bottom-up~\cite{hua2005learning} approaches have been proposed. The research community also retrieved the spatial configuration of humans by matching the holistic human shape~\cite{mori2002estimating}, aggregating poses from segmentations, or using contours to model the shape of human body parts~\cite{kumar2009efficient}. Recently, many new ML models are introduced that can detect and connect human body parts to detect human pose \cite{andriluka2018posetrack}, \cite{cao2017realtime}. These approaches suffer from the huge computation burden of image analysis for each and every object in the frame; and consequently, the computing time may be longer than one second in a crowded scene \cite{penmetsa2014autonomous}. Besides the long delay to recognize the pose of an object, the detected pose is to be compared to predefined ``normal poses'' to detect anomalous ones, such that the detection accuracy highly depends on the quality of the training set. If the training set fails to provide a sufficiently large number of ``normal pose'' exemplars, the system could suffer a high false positive rate. To reduce the need for a complete training data set, we seek a model to estimate between representative CNN extracted examples using fuzzy logic.

\subsection{Fuzzy Controller}

Fuzzy sets and fuzzy logic together make fuzzy controllers – which is an attractive and promising method \cite{karaboga2018adaptive} to deal with uncertainties. While the concept of fuzzy mathematics and its use is not new, it is specifically useful when methodologies are too complex for analysis by conventional quantitative techniques or when the available sources of information are interpreted in a qualitatively inaccurate or uncertain way \cite{yen1999fuzzy}.

Fuzzy logic, which is the formal operator on which fuzzy logic control is based upon, emulates human thinking and natural language process rather than the traditional Boolean logic \cite{lee1990fuzzy}. It is an effective tool for capturing the approximate, inexact nature of the real world. Viewed as a semantic classifier, the underlying part of the fuzzy logic controller (FLC) is a set of linguistic rules related by the dual concepts of fuzzy implication and the compositional rule of inference. The FLC provides the automation technique which can convert the linguistic control strategy based on expert knowledge into a stand-alone control system that is robust to environment and measurement noise \cite{yagishita1985application}. When the FLC is applied to domains for which it is trained, reports provide evidence of FLC superior results when compared to those obtained by conventional control methods. Hence, fuzzy logic control is considered as a step toward bridging conventional precise mathematical control to human-like decision making \cite{gupta1980fuzzy}. 

Because of the high uncertainty and complexity in describing the human activities, fuzzy decision making is appealing to address suspicious human activity detection in public safety surveillance systems.

A comprehensive survey for video classification and anomaly detection for video analytics was given in \cite{lim2015fuzzy}. Some works are focused on specific human behavior classification \cite{yao2015fuzzy}. Both methods used the context of the data to show human decision making and behavior prediction, instead of anomalous behavior. Researchers have also tried to reach better accuracy for human behavior prediction by combining the fuzzy logic and a Hidden Markov Model (HMM) \cite{mozafari2011novel}.


\section{I-SAFE System Architecture}
\label{sys}

\begin{figure}[t]
    \centering
        \includegraphics[width=0.5\textwidth]{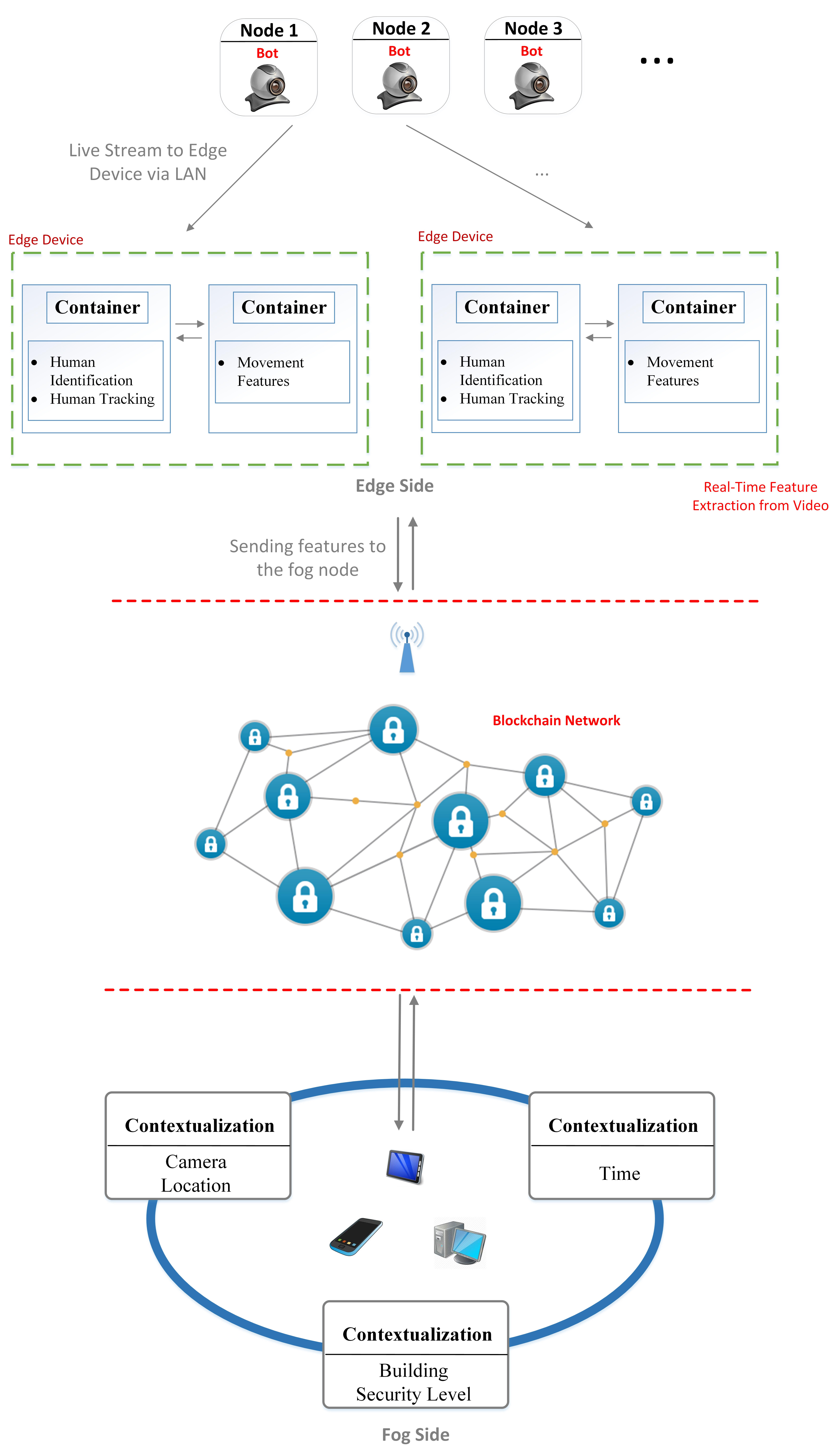}
    \caption{The smart surveillance system architecture for I-SAFE scheme.}
    \label{fig:structure}
    \vspace{-10pt}
\end{figure}

In a well performing safety system, humans are detected and their movement and behavior is closely watched to draw a conclusion of their intent. This process may be performed by the human agents or a smart system. The proposed I-SAFE system tries to mimic the same decision-making logic as the human agent. Figure \ref{fig:structure} presents the decentralized smart surveillance system architecture proposed to enable the I-SAFE. Within the architecture, the functionality of I-SAFE scheme can be considered as three steps:

\begin{itemize}
   \item \emph{Video Feature Extraction}: Based on the design of the decision making algorithm, features from the video are extracted. These features present the movement of the humans as objects of interest in the frame. A lightweight CNN that feeds an online tracker, gives the position of the objects in each processed frame. In finding the domain-selected features for decision making, I-SAFE calculates the relative speed of each human in the frame and also the relative movement direction. 
   
   \item \emph{Feature Contextualization}: Obviously, the context of the data can support the decision. Experiences from law-enforcement officers confirm that time and geo-location are factors which have an undeniable affect on the outcome whether an officer engages in for further investigation. Realizing this importance, I-SAFE incorporates contextual information to the features before feeding them to the decision making algorithm.
   
   \item \emph{Fuzzy Decision Making}: As one of the more robust approaches to control an environment, a fuzzy system armed with a complete set of rules and comprehensive membership functions (please refer to Section \ref{fuzzy} for details) can make a powerful tool for decision making. The fuzzy relations between inputs and outputs enhances system robustness to noise. This effect is amplified by the object tracker and the feature extraction imperfections that lead to added noise to position-related features. 

\end{itemize}

As illustrated in Fig. \ref{fig:structure}, the I-SAFE system utilizes edge camera units. The edge device is either a basic surveillance camera with a Single Board Computer (SBC) mounted, like a Raspberry Pi board, or a smart camera with integrated computing resources. The edge node processes the video frames and extracts the features for decision making purposes \cite{nikouei2018kerman}. Under the framework of I-SAFE, although the live feed is available for the human operators in the control room, it is not recommended due to the high traffic volume, and it is impractical to expect any human operator to identify useful information from hundreds of real-time video streams. Only the extracted features that are essential for decision making are immediately transferred to guarantee low network communication workload. Exploiting a web service, the camera video stream can be stopped in case of no requests. Additionally, data context is not added to features at the edge, because of the repetition that is involved leading to undesired overhead in feature communication. 

The feature stream is outsourced to a fog device located closest. A fog node can be a smart-phone, laptop, or a desktop, which is more powerful than the edge device and is deployed near the source of data. The reason for this outsourcing is simply the limitations of the edge node. After video processing, more calculations could drastically impact performance as verified by our experimental study reported in Section \ref{experimental}.

Adopting the \emph{decentralized approach}, the I-SAFE scheme possesses several advantages over the traditional cloud-based services. In terms of the system architecture design, the network manager is eliminated that usually becomes a bottleneck of performance as the number of nodes increases. According to the capacity of a fog node, a number of edge servers are assigned to it at system setup. And the operator may have access to the real-time stream from the edge or the decisions made by the fog services.

Figure \ref{fig:structure} shows that the access to the edge is managed by a private \emph{blockchain access control} (BAC) protocol. The access authentication is conducted in network setup phase and the smart contract is enforced in the blockchain network. To ensure the whole platform is scalable and easy-to-upgrade, both the video processing and security management functions are implemented using microservices architecture. Each microservice is placed inside of a docker container with all requirements necessary for ease of distribution. Due to the limited space, the rationales, architecture, implementation, and the performance evaluation are not presented in this paper. Interested readers are referred to two papers for more information, one details the microservices based surveillance platform \cite{nikouei2019decentralized} and the other is focused on the blockchain enabled security mechanisms \cite{xu2019blendmas}.

\section{Feature Preparations}
\label{data}

In the I-SAFE scheme, dynamic data is analyzed to prepare features for the fuzzy engine with the following steps: feature generation, selection, extraction, and contextualization, where the classifier works best if the features produce most divergence between classes. Deep Learning (DL) models, which auto-define the features, need labeled training dataset that are unfortunately not available for security use-case. Thus, selected features based on subject matter expert (SME) are presented in this section. Moreover, context supports robust operations by including environmental, societal, and cultural information. 

\subsection{Feature Selection}

Looking at recent publications for anomaly detection, there are many different features that could be of importance. However, this work utilizes law enforcement officers to have a better understanding of what features should be of interest. Based on their input, loitering in odd hours or places, indicates high chance of misbehavior. Although there are other clues such as the appearance, clothing, and certain smells that may also draw an officer attention, in order to minimize bias or profiling individuals and avoiding extracting sensitive private information from the individual, the system only considers the pattern of movement after tracking the general figure of a human body. In the future, gesture and important body keypoints can be added to the tracking module for more accurate description of the activities. 

The pattern of individual's movement is an important factor in the decision making procedure of an officer. Actually, many ML models can learn from the dynamic time-series data to model patterns and detect the anomalies \cite{aved2015multi}. In a 2-D RGB picture of a random scenario, the pattern of movement may be confusing as depth information is also presented in up or down movements. In addition, studies show that movement patterns are context specific \cite{zhang2017video}. For example, a person who is walking to his/her car in a parking lot may behave differently from a person who is walking into his/her dorm room. Thus, comparing the patterns and directions generalize scenarios leads to high false positive rate. Furthermore, mapping the pattern of movement is time consuming and resource demanding. Thus, it is not affordable for resource limited edge or fog devices for each person. Instead, indicators of the movement are chosen because of their generalization to all scenes. 

A strategy is employed to utilize the number of speed and direction changes, in order to have the moving pattern. The more number of changes indicates the higher probability of loitering. Again according to the law officers who watched pedestrian movements closely on a daily basis, a person who has a known destination in mind is likely to walk straight an at certain reasonable pace. Turning around in an area or changing position without an apparent destination should raise an alarm. The other benefit of this strategy is that there is no need for extracting complex pastern routes for each person, rather the indicators. The difference in calculation time is more noticeable when there are more than three objects are present in the frame. 

There are two complimentary features. \emph{Standing at one location for a long time} may be an indicator of loitering. Nonetheless, this feature should be used in context. In addition, if there are more than one person in the scene, it is less likely to be of suspicious. Thus, the \emph{number of the people} in the frame is also considered as a feature.

\subsection{Video Feature Extraction}

The human detection uses a Convolutional Neural Network (CNN). If a human as the object of interest is detected, the object's bounding box coordination resides in a queue. In each upcoming frame, the queue is updated with the tracker bounding box prediction. Once every several frame when the CNN is applied to the input frame, not only it will add any newly detected objects to the queue for tracking, but it also checks the object placement in each previous bounding box. If the Intersection Over Union (IOU) is smaller than a threshold which is set by the administrator, it will add the person as a new object and delete the old bounding box and information related to it. 

\begin{figure}[t]
    \centering
        \includegraphics[width=0.48\textwidth]{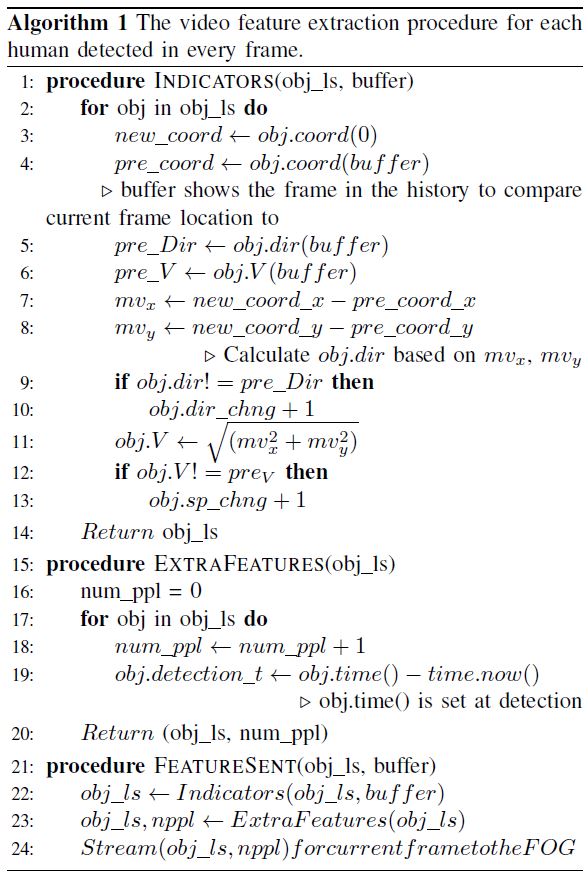}
    \caption{Pseudocode of Algorithm 1.}
    \label{fig:algo_1}
    \vspace{-10pt}
\end{figure}

The online tracker should give an accurate estimation of the position of each object, otherwise the extracted features lead to inaccurate classification. The Kerman hybrid tracker \cite{nikouei2018kerman} maintains tracking and supports track hand-off to improve tracking accuracy. By extracting the coordination of the object in the frame and comparing it to the previously collected information, we can obtain the indicators and features for the decision making. Post processing algorithm of the frame after reception of each bounding box is presented in Fig. \ref{fig:algo_1}. The feature set for each frame is then transmitted to the fog unit to be contextualized and change detection determination. 

It should be mentioned that the accuracy of the I-SAFE system incorporates algorithms that detect and track the human as the object of interest. Although the algorithms used for human detection and tracking at the edge utilize deep learning architectures and methods, the performance accuracy is not 100\% which means that the camera may lose the object of interest and so the features used in the fog node will be disrupted. The I-SAFE framework has a disunified architecture where the performance can be improved by integrating new algorithms.

\subsection{Feature Contextualization}

The first step toward a dynamic data analytics is to consider all of the features that explain the input data including feature generation method. Feature generation is learned from previous analysis with salient feature selection. In case of human-oriented public safety surveillance, features are directly extracted from the video frames (e.g., intensities, lines, shapes) as well as other external factors (e.g., camera placement, lighting conditions, and scene content). The procedure of putting these features together with the factors will be referred to as \emph{feature contextualization}.

This paper focuses on university campus surveillance and attempts generalization. During the normal operating hours of a building on campus; many students, faculty, staff, and other personnel may be detected in the scene. While it is normal to observe many people during the day, it is abnormal if many people appear after 11:30 pm. In the case of the abnormality detection, the contextual data of the time of day assists in decision making. According to the police, the time of day determines officer vigilance in monitoring attention to a gathering. What context features should be selected and how common they are utilized in cases of surveillance are determined based on the suggestions from our campus police officers.  
The edge device that hosts the data extraction from the video and prepares the feature list for each frame, cannot handle the contextualization of features due to the resource constraints. The video processing task makes use of most of the computational power \cite{nikouei2018realtime} while the rest is allocated to the transmission and security modules. Moreover, the context, such as spatiotemporal and geo-location are usually repeated data that sending them with each frame creates overhead. Therefore, the video features are extracted by the edge camera and sent to the fog node for contextualization. There are three features that are added to the camera data during the contextualization phase. 

The first is the \emph{time of the day}. The importance of this feature may differ from one location to the other, but it is considered as one of the most important factors to make a decision in security systems. The second factor is the \emph{geo-location of the camera}. Cameras installed indoors should have a different set of thresholds for decision making than the outdoor cameras. Just as the accessibility and space use-case varies, normal behavior changes too. Additionally, cameras that are installed outside of a bank should have less tolerance for detection of a human being after hours and should raise an immediate alarm in such a case. Thus, the \emph{security level} of the building where the camera is mounted is the third context we consider. 

\section{Fuzzy Model}
\label{fuzzy}

\subsection{Rationale}

The decision-making process for the safety surveillance system is based on a fuzzy control system model. Although the momentum in FLCs systems is lost due to absence of experts in many challenges, the fuzzy method remains one of the best methods for systems with high noise levels. In surveillance, the law officers can act as the experts and their opinion shall be used for system operation. The officers take months or even years to develop an innate sense of behavior analysis for a certain location of their duty. 

A DNN training that can do classification with proper accuracy requires many training sets both negative and positive examples and while negative data is easy to acquire, positive samples are harder to gather. Even after labeling, if the dataset does not contain all scenarios, the result does not cover the whole input space which leads to undetected events.

In case of the campus safety system, there are many campus police officers who have spent time in the domain of the interest and they know what to expect from the crowd. Their experience is used for creating a series of rules that are implemented in the fuzzy model to detect the anomalies. On the other hand, general purpose classifiers with unsupervised learning methods do not offer high accuracy and suffer form noise distortion, trivial solutions, and collapsing of features in deeper models \cite{bojanowski2017unsupervised}.

With the contextualized features, a fuzzy control system is introduced to mimic how the police officers make decisions and to obtain a semantic output amenable to human operators. Unlike mathematical probability analysis, the fuzzy-based models are not based on numbers, but \emph{semantic classes}. A FLC maps the input sensor measurements to linguistic labels, which are a description of the input. The fuzzy system affords the mapping of operator knowledge into a decision making model.

For anomalous activity detection, the officer performs linguistic-type reasoning in his/her mind and reaches a conclusion of a behavior being normal or abnormal, instead of giving a numeric description of the observation. To mimic this cognitive behavior, the fuzzy model gives a linguistic output which is the classification label. This output can be translated to a number based on the defuzzification formulation that is consumed by digital computers. The results are reported to the police department with the amount of attention (i.e., based on a confidence, credibility, or reliability estimate) needed to assess a specific scene. 

\subsection{Fuzzy Model}

The feature-set for each of the objects of interest is sent to the fuzzy logic controller at the fog node where it contextualize and fed the FLC. 

The first step to realizing the fuzzy model is fuzzification, which translates the features to a fuzzy value. For any set $X$, membership functions represent the fuzzy subsets of it. For an element $x$, the fuzzy subset $A$ corresponding value is denoted by $\mu_A(x)$ as shown as Eq. (\ref{eq:A}):

\begin{equation}
 A \{x , \mu_A(x) ; x \in X\}
 \label{eq:A}
\end{equation}

In order to fuzzify the measurements (the contextualized features in this case), each measurement is compared to its respected subsets $\mu_{A_i}(x)$ as shown in Eq. (\ref{eq:w}), which creates the linguistic variables that are used in the rules. Linguistic variables are in contrast to normal variables where each variable presents a range of meanings such as {cold, medium, hot}. 

\begin{equation}
 w^x_i \max_x[\min(A^\prime(x), \mu_{A_i}(x))]
 \label{eq:w}
\end{equation}

\noindent where each measurement is considered as a range shown by $A^{\prime}(x)$. If $x$ is only one number as the measurement, then $A^\prime(x)$ becomes only one number too. Note, subscript $i$ represents each of the fuzzifiers (membership functions).

If two sets $X$ and $Y$ are considered in a fuzzy system, based on each fuzzifier in each set for each reading, linguistic variables $w^x_i$ and $w^y_i$ are calculated. The linguistic variables are used in the rule set to calculate the results ($C^\prime_i(z)$) for each rule. The Minimum value between the resulting Premise $W_i$ and each fuzzy membership functions for the output $C_i(z)$ yields Eq. (\ref{eq:rule}):

\begin{align}
 W_i  \min[w^x_i, w^y_i] (AND_{Operand})\nonumber\\
 W_i  \max[w^x_i, w^y_i] (OR_{Operand})\nonumber\\
 C^\prime_i(z) = \min[W_i, C_i(z)]
 \label{eq:rule}
\end{align}

\begin{figure*}[t]
    \centering
        \includegraphics[width=0.98\textwidth]{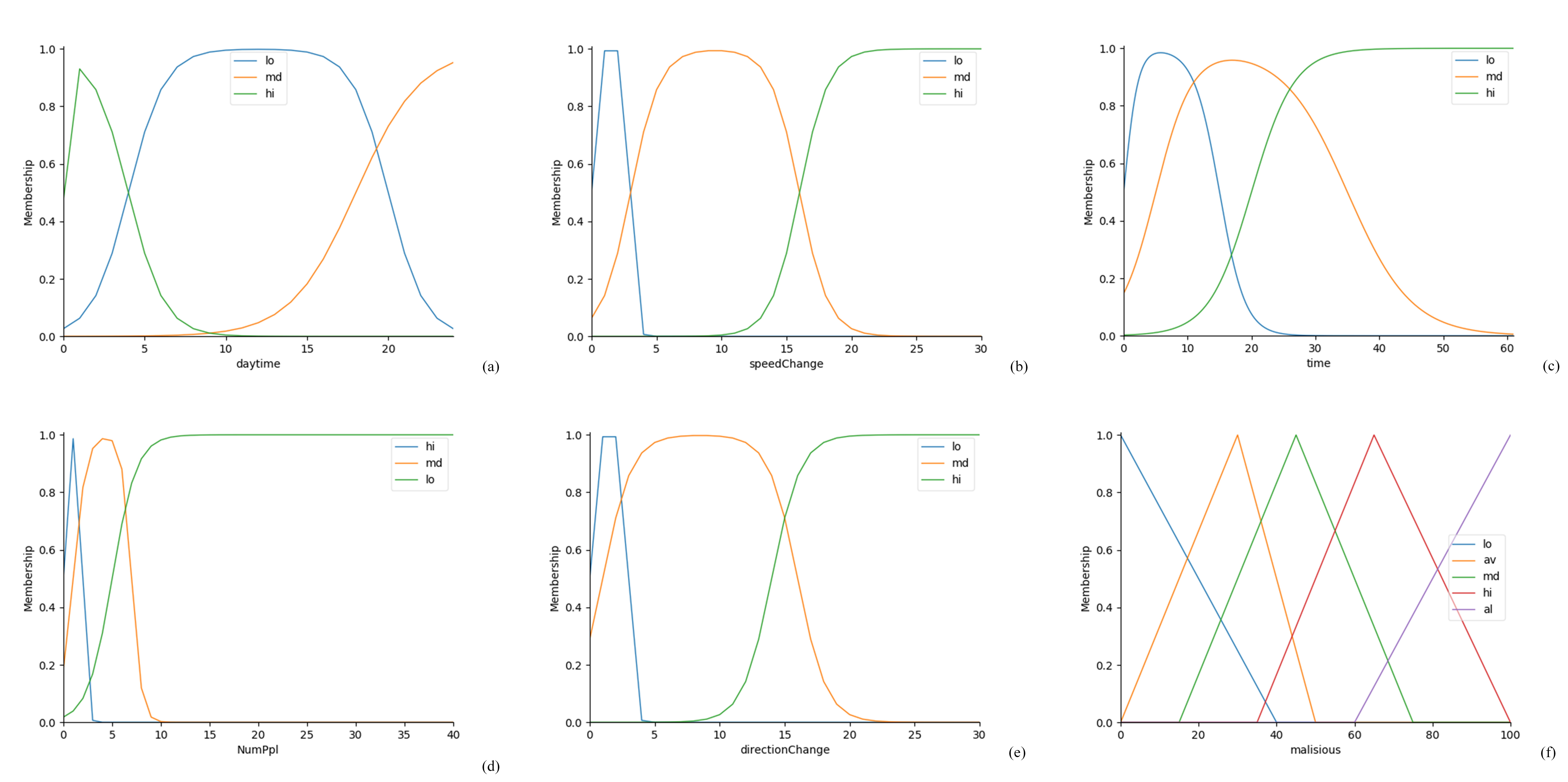}
    \caption{Membership functions: (a) The hour in the day. (b) The number of times there is a change in the speed of the object. (c) The total time that the object is in the frame. (d) The number of people present in the frame. (e) The number of times there is a change in the direction of movement of each object. (f) The malicious behaviour levels.}
    \label{fig:memfunc}
    \vspace{-10pt}
\end{figure*}

\subsection{Membership Functions}

According to Eq. (\ref{eq:w}), the linguistic variable is mapped to an interval [0, 1] which can be inferred as a credibility analysis. If the credibility of a subset is less than $0.5$, then the linguistic variable is not reliable enough. This yields to subsets (membership functions) that should cover each set so that at no point $x$, the credibility of aligning a feature to a set falls below $0.5$. This in return yields that the output results of the system can have higher credibility as the inputs are more reliably fuzzified. In addition, it is very important to consider the best membership functions to cover the entire set. Having the wrong candidate as the membership function can lower the credibility of that set faster or the membership function will not cover the desired area in the set. 

Figure \ref{fig:memfunc} presents the fuzzy-relation types, where the first five sets (Fig. \ref{fig:memfunc}a - e) are the input membership functions and the last one (f) is the output. 

In Fig. \ref{fig:memfunc} the x-axis shows variable $x$ in \ref{eq:A} and the y-axis represents the credibility of membership functions that $x$ belongs to as it changes. In part (d) for example, having five people in the frame, means about $1.0$ confidence in ``medium activity'' and $0.27$ confidence in ``high activity''. Then Eq. (\ref{eq:w}) determines that having five people in the frame means ``medium activity'' and it is considered for linguistic variable of set: ``NumPpl''.

Three membership functions are considered for each feature interval and five membership functions are used to describe all possible levels of suspicious behavior with high accuracy. As illustrated in Fig. \ref{fig:memfunc}, the following contextualized features are presented to the fuzzy system, thus the input is fuzzified and the output is made based on rules: 

\begin{itemize}
    \item (a) The time of the day, which is the hour ranging from [0:00 to 24:00];
    \item (b) The number of the times an object in a frame changes the speed [0 30]. If a person walks for a long time in the frame and changes directions many times showing not having a clear destination, there is a good chance of loitering;
    \item (c) The time that a human object stays in a frame, normally a person walks out of the frame in several seconds if they are walking at normal speed [0 30] seconds;
    \item (d) The number of people in the frame under processing [0 40]; and 
    \item (e) The number of the times an object in a frame changes the direction [0 30].
\end{itemize}

The boundaries of each set are designed to handle almost every possible scenario, but in case of outliers, the fuzzy system can handle noise and out of range values very well. The key is the fuzzification process. During the calculation phase, if the values are outliers (out of the set scope), the fuzzification still maps the linguistic variable that is most closest to the edge of the set limit. If the fuzzy system fails to align measurements to set values, the human operator will receive an acknowledgement that the inputs are erroneous and alert the operator to observe the situation in the videos. Reported in an error log, the output is also set to zero so no alarm is raised.   

It is noticeable that in the models of Fig. \ref{fig:memfunc}, each set is covered with at least one of the membership functions for any given input, and there is no point on the x-axis that at least one membership function is above $0.5$ of the nominal value. Which implies that the decisions of set classification are made with high credibility. 

Having a membership function that has a salient value for a long range translates to higher share and higher power of corresponding linguistic variable. All membership function shapes are chosen keeping this in mind. And is why in the malicious behavior membership functions just one point is the highest confidence in that result so no bias toward one behavior is not added. 

In a completely defined fuzzy control system, all possible combinations of the fuzzy linguistic variables should be considered in the rule set. However, there is a exponential relationship between the number of rules and the processing time. Hence, the rule set for I-SAFE is designed to cover all scenarios of interest while utilizing a combination of the features.

\subsection{Producing the Output}

\begin{figure*}[t]
    \centering
        \includegraphics[width=0.7\textwidth]{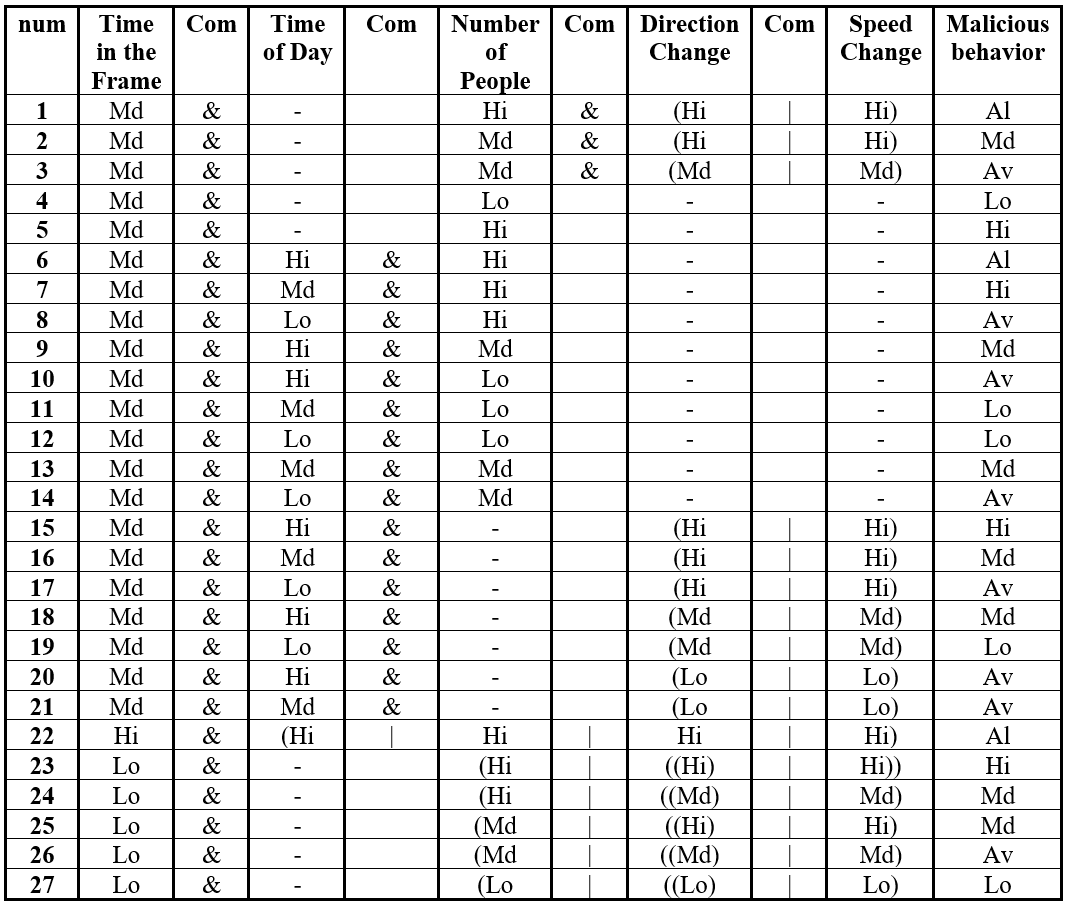}
    \caption{Set of rules that are used for the safety system.}
    \label{fig:rules}
    \vspace{-10pt}
\end{figure*}

Figure \ref{fig:rules} shows how the features are combined in rules to reach a conclusion. The $Com$ column shows the combination between the linguistic conditions (features). Only logical Intersection ($AND$) and Union ($OR$) are used, and the parenthesis show which logical operation should be executed first between the conditions. Since the speed change and direction change variables are subject to high measurement noise, the I-SAFE system employs the OR operator between them to reduce their impact based on the number of rules using them.

Based on the five linguistic variables respective to each fuzzy set for every contextualized features, the system gives the suspicious activity fuzzy probability. After defuzzification, the geo-location of the camera and building security level can be considered to give an appropriate threshold to raise the alarm. Such that the camera position and building security are considered as the two final features.

Once again, the expert's experience or knowledge enforces the rules to control the environment. The rules may be different corresponding to different conditions. The expert is asked to generalize rules and features that are vital in the decision-making process. Note that each membership function boundary that is used in the input and output generations in Fig. \ref{fig:memfunc} is based on a camera that is installed on a hallway in a campus building. If the area under security surveillance needs more supervision, the administrator can change the fuzzifiers subject to tolerate less activity and/or raise the alarm sooner in certain times. All changes are done at setup and no more adjustments are needed.

The rule set used in I-SAFE system emulates the calculations that an officer performs before approaching a suspicious object. As shown in the fuzzy sets, the number of people in a frame is an important feature. The highest attention is drawn to scenarios where only few people are present at night. Based on the law enforcement experience, the videos that have only one person in frames that are the most important. Therefore, the video footage with one or two people are considered as of high interest. With increase in the number, the interest goes down based on the double sign membership functions. Another key area is the amount of time that the object is present in the location. As the time expands, the probability that the object is loitering goes higher with respect to its set membership functions.

\subsection{Suspicious Score}

The last step is to defuzzify the results of the rules and translate it to a number between 0 to 100\%. A threshold decides whether or not the output should raise an alarm to make operators aware of an activity. Then the operator will make decision for further actions. The $C^\prime_i(z)$ can be defuzzified using Eq. ( \ref{eq:defuzzy}):

\begin{equation}
 F^{-1}(c) = \frac{\int (W_i\mu_c(x)) }{\int \mu_c(x)}
 \label{eq:defuzzy}
\end{equation}

\noindent where the $\mu_c$ is the membership function of the output. 

The I-SAFE system is able to draw attention to the scene where anomalous or suspicious activities are determined, but it is the human operator that makes the decision of action. The fuzzy model is implemented on the fog level devices and it is easy to access and reconfigure parameters through a single cloud node, for a batch of edge units connected to the fog, if the operator chooses so.

\section{Experimental Results}
\label{experimental}

A proof-of-concept prototype of the I-SAFE scheme has been implemented and tested using real-world surveillance video streams. The experimental results are encouraging that the design goals are achieved to provide a secure, agile, and fast surveillance system for safety monitoring. The I-SAFE system detects the activity successfully in an average of 0.002 seconds after features are pre-processed.

\subsection{System Setup}
The prototype consists of both edge and fog layer function units. At the edge, human detection and tracking are accomplished using lightweight L-CNN and Kerman algorithm. Features are created using the indicators and other methods explained previously. The edge functions are hosted by a Tinker Board with 1.8 GHz ARM-based RK3288 SoC and 2 GB LPDDR3 dual-channel. The Tinker Board is placed behind the camera, in this sense, the camera can be considered as the sensor and the edge device is the Tinker that connects to the sensor through Local Area Network (LAN). The features are sent to a laptop PC running Ubuntu 16.04 operating system as a fog node, where the contextualization and fuzzy decision making of the I-SAFE scheme is located. The PC has a 7th generation Intel core i7 processor @3.1GHz and 32 GB of RAM. The wireless connection between the fog and edge is through wireless LAN (WLAN) with 100Mbits/s. 

\subsection{Threshold Setting}

\begin{figure}[t]
    \centering
        \includegraphics[width=0.425\textwidth]{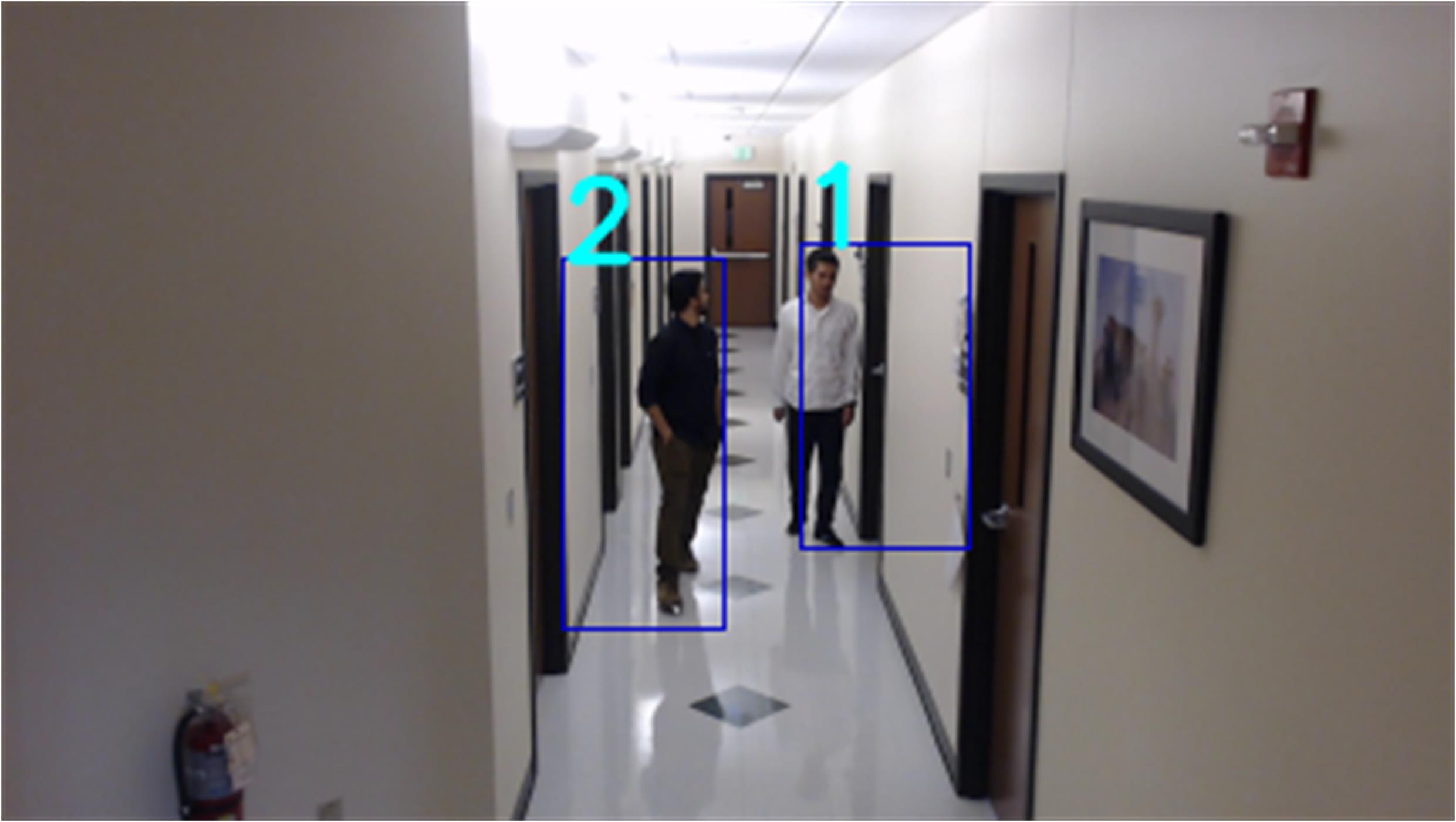}
    \caption{Two objects normally walking in the frame.}
    \label{fig:sidebyside}
    \vspace{-10pt}
\end{figure}

Suspicious activity can be interpreted in different ways; but the goal was to provide a machine-level triage of the situation for cueing operators to human abnormal behavior. For example, the abnormal behavior could point to a bicyclist in a crowded place where everyone else is walking \cite{xu2015learning}. In another attempt a certain possible pose of a human is a sign of abnormal behavior \cite{penmetsa2014autonomous}. The challenge is to determine an ontology of activities that would alert and operator to potential abnormal activity. In this paper, we consider a campus environment, where simply put, the students are unlikely to loiter around in parking lots or in hallways in late hours. The system is designed in such a way that it will require certain thresholds to be met before raising an alarm.

It is important to note the separation of the feature map and decision making algorithm, makes the project more suitable for edge computing paradigm where outsourcing the process to higher link in the hierarchy is indisputable.

Figure \ref{fig:sidebyside} is a scenario where two people are walking at their normal pace and they will exit the frame when reaching the end of the hallway. The algorithm follows both objects and outputs the abnormality score corresponding to the measured likelihood of suspicious activity respective to each individual.

\begin{figure}[t]
    \centering
        \includegraphics[width=0.5\textwidth]{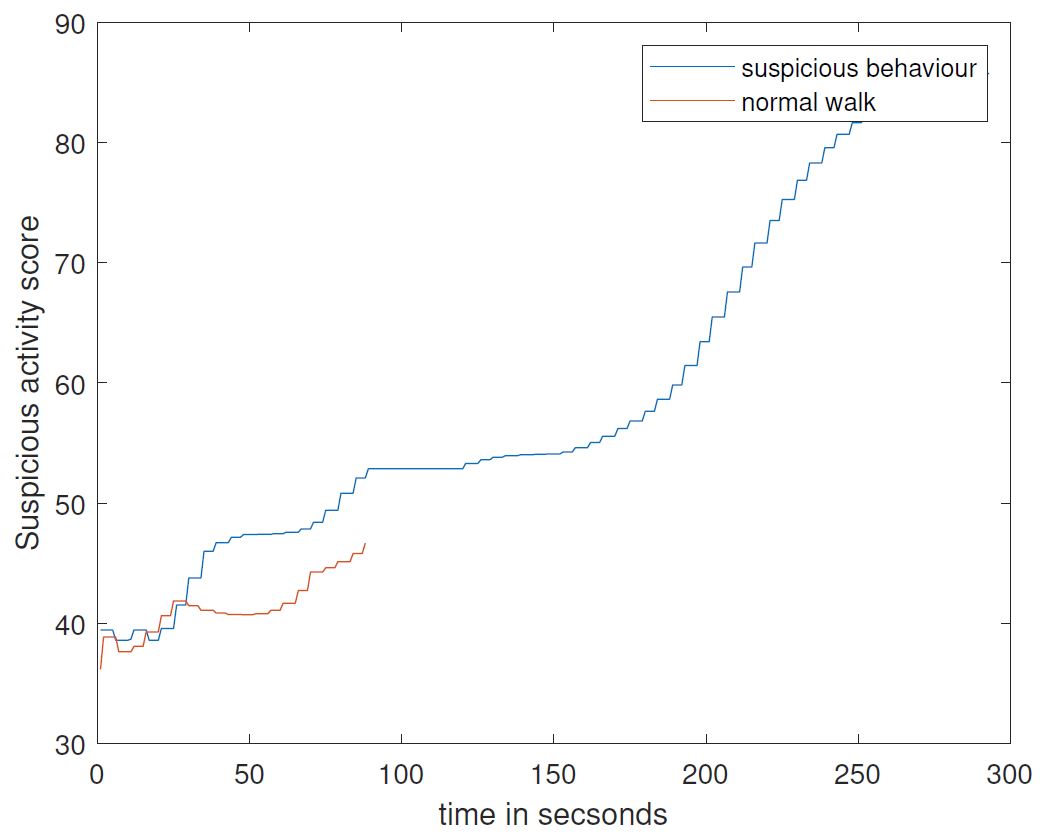}
    \caption{Comparison of two sample cases in decision making process.}
    \label{fig:numbers}
    \vspace{-10pt}
\end{figure}

Figure \ref{fig:numbers} compares the malicious scores of these cases. The x-axis is the time an object in the frame in second, and the y-axis is the defuzzified suspicious score. The red line is the score of the case of single person walking at 11:00 am, where the object walked through the corridor in about 100 seconds. The score is in a reasonable range showing no suspicious behavior. In contract, the blue curve is the score of the single person walking at 3:00 am and stays in the area for long time. The suspicious score is rising as time goes by as the blue curve starts at a higher value at time zero than the red curve does, because the scheme considers walking at 3:00 am is more suspicious. As the time of movement increases, the normal activity's score rises slower because of the other parameters that besides the time shows suspicious activity, however the blue line has higher jumps in the score.

Figure \ref{fig:numbers} does not indicate the threshold to raise the alarm. It can be set conveniently by the system administrator based on the experience of building usage. In another words, setting a reasonable threshold needs a statistical analysis of the distribution of activity scores corresponding to the behavior patterns in this building.

\subsection{Performance Evaluation}

\begin{figure}[b]
    \vspace{-10pt}
    \centering
        \includegraphics[width=0.41\textwidth]{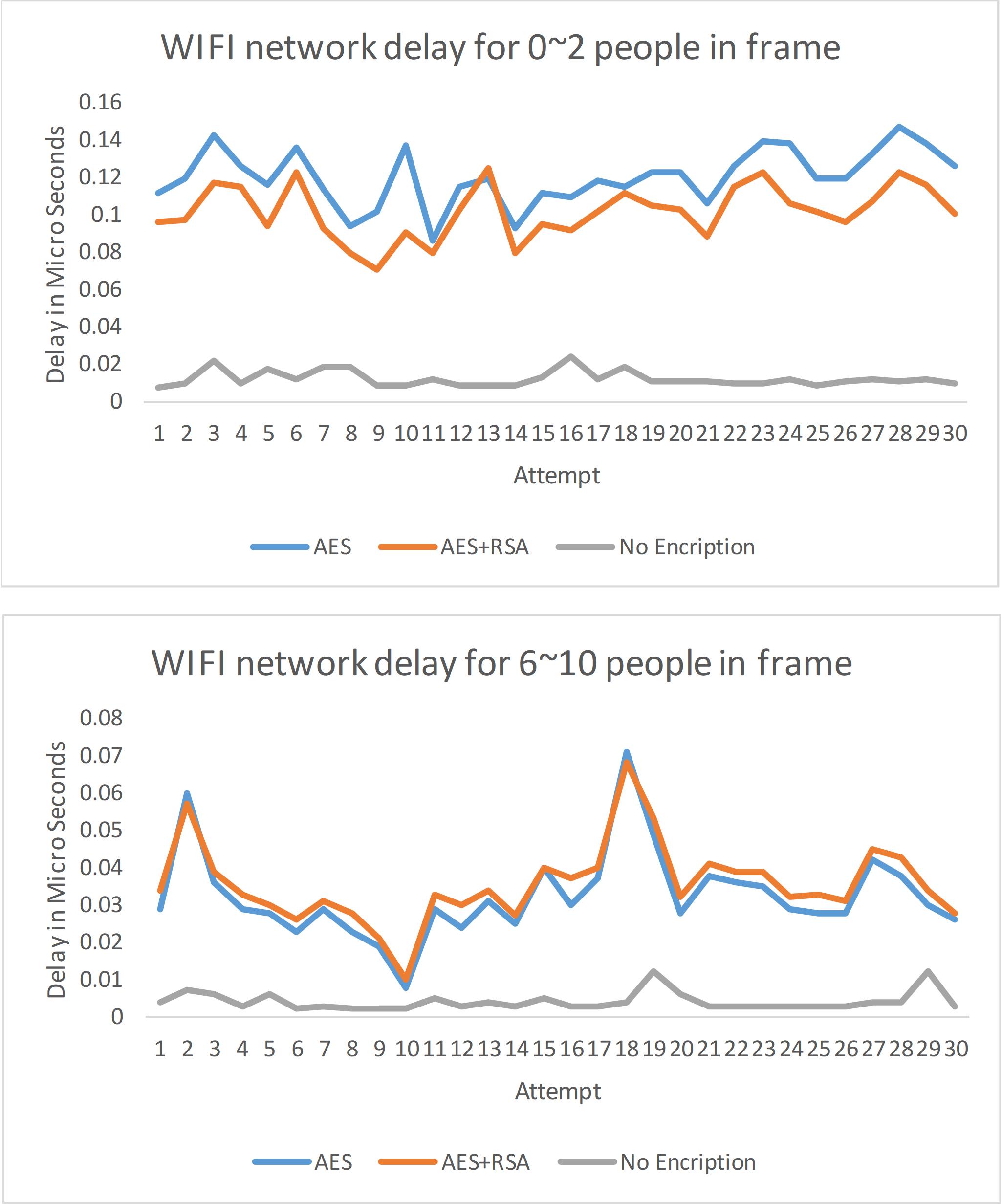}
    \caption{Time delays (in micro-seconds) as the result of the data encryption and transmission on the local wifi network.}
    \label{fig:delay}

\end{figure}

Figure \ref{fig:delay} shows the delay due to the data transfer from edge to the fog. Three scenarios are compared: no encryption, AES (Advanced Encryption Standard) encryption, and AES+RSA (Rivest-Shamir-Adleman) encryption. The AES+RSA has handshake and establishment of connection based on the RSA and the rest of the data transaction is based on AES, which is better in terms of low latency on resource limited devices. It is noticeable that the delay does not have substantial impact to the real-time performance of the system. Since five to eight frames per second is the speed at which the edge device can process the input frames. In addition, with the increment of the number of detected human objects, the feature file being transmitted for each frame gets bigger and also the communication takes more time. As more data is transmitted there are spikes in the transmission times due to unstable network connection. Two scenarios are included in Fig. \ref{fig:delay}, one where there are between 0 to 2 objects in the frame and another where we have 6 to 10 objects in each frame for comparison of delay as the files get bigger.

\begin{figure}[t]
    \centering
        \includegraphics[width=0.46\textwidth]{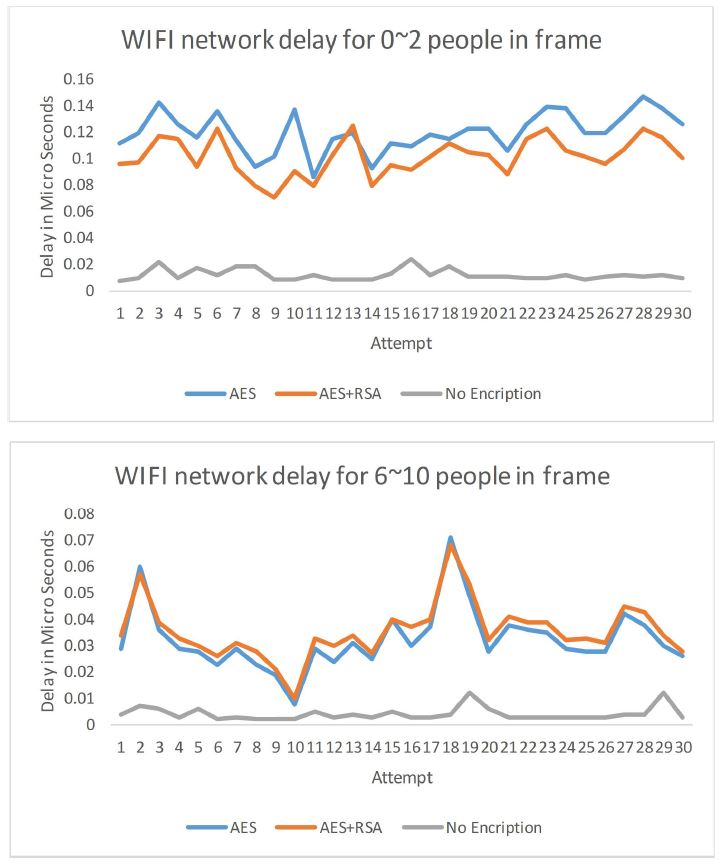}
    \caption{Time (in micro-seconds) needed for decision making given the features from the video.}
    \label{fig:delayfuzzy}
    \vspace{-10pt}
\end{figure}

Figure \ref{fig:delayfuzzy} compares the difference between processing the fuzzy model at the edge or at the fog level. The total time shown in Fig. \ref{fig:delayfuzzy} includes the time for data contextualization and the fuzzy control system results. As shown in Fig. \ref{fig:delay}, the communication time is much shorter than the time required in decision making process. So the communication time is neglected in Fig. \ref{fig:delayfuzzy}. Note how the edge devices struggles to have around 1.5 (FPS). The same operation takes about 0.002 second on the fog node. Figure \ref{fig:delayfuzzy} justifies outsourcing the fuzzy decision making function to the fog node along with showing the heavy load on the edge devices. Processing a frame and generating a decision in 0.002  seconds in average for human activity detection meets real-time requirements. Considering the velocity of pedestrians, a person cannot move much in 0.1 seconds (processing rate of the current video processing applied at edge) giving ample time for a security response. 

In the future, with the introduction of more powerful edge devices, the whole process may be executed at the edge. Figure \ref{fig:delayfuzzy} is generated based on a scenario of two people in the frame. Intuitively, if more people are in the image frame, longer delay is expected both for video processing and decision making.

\begin{figure}[t]
    \centering
        \includegraphics[width=0.46\textwidth]{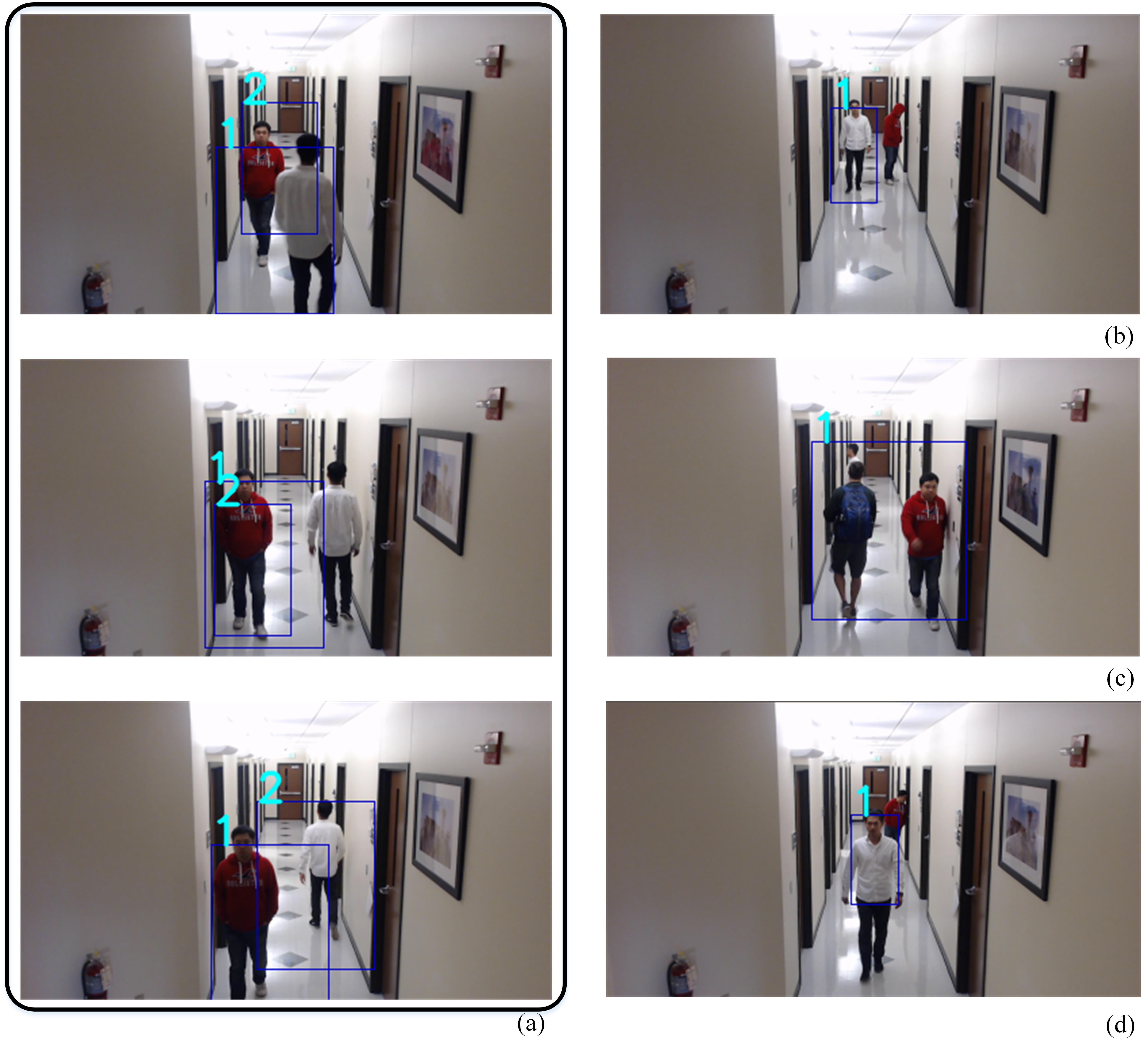}
    \caption{Moments of the human detection and tracking failure that makes the decision making harder: (a) where the tracker loses the object and finds him again. (b) Object is too small for the detection algorithm to detect. (c) The detection confuses two people as one. (d) One object covers the other so he/she will remain undetected.}
    \label{fig:fails}
    \vspace{-10pt}
\end{figure}

Figure \ref{fig:fails} presents some occasions where the detection and tracking algorithms failed. As explained before, this leads to a lower accuracy of the decision making. Figure \ref{fig:fails} includes three instances. Part (a) of this figure shows the object of interest (person in white) at the beginning, who is closer to the camera and is walking past the other person. The tracking algorithm stops following the person in white and stays with the person in red as the red one becomes closer. The detection algorithm, however, detects the person in white again and deletes the other bounding box. Unfortunately, the suspicious score data gathered for the person in the white shirt is lost. Figure \ref{fig:fails}(b) shows a scenario where the person in red is far away from the camera and the system failed to detect them. Figure \ref{fig:fails}(c) is when the detection algorithm detects only one person instead of two that are in the frame. While this problems exists, it happens very seldom over numerous trails. Finally, Fig. \ref{fig:fails}(d) shows a very challenging case, when one object blocks the other, as there is no way to have both two people detected. Each of these issues can be mitigated through intelligent design such as more cameras and enhanced trackers, of which the general operation is scalable. 

\begin{figure}[t]
    \centering
        \includegraphics[width=0.46\textwidth]{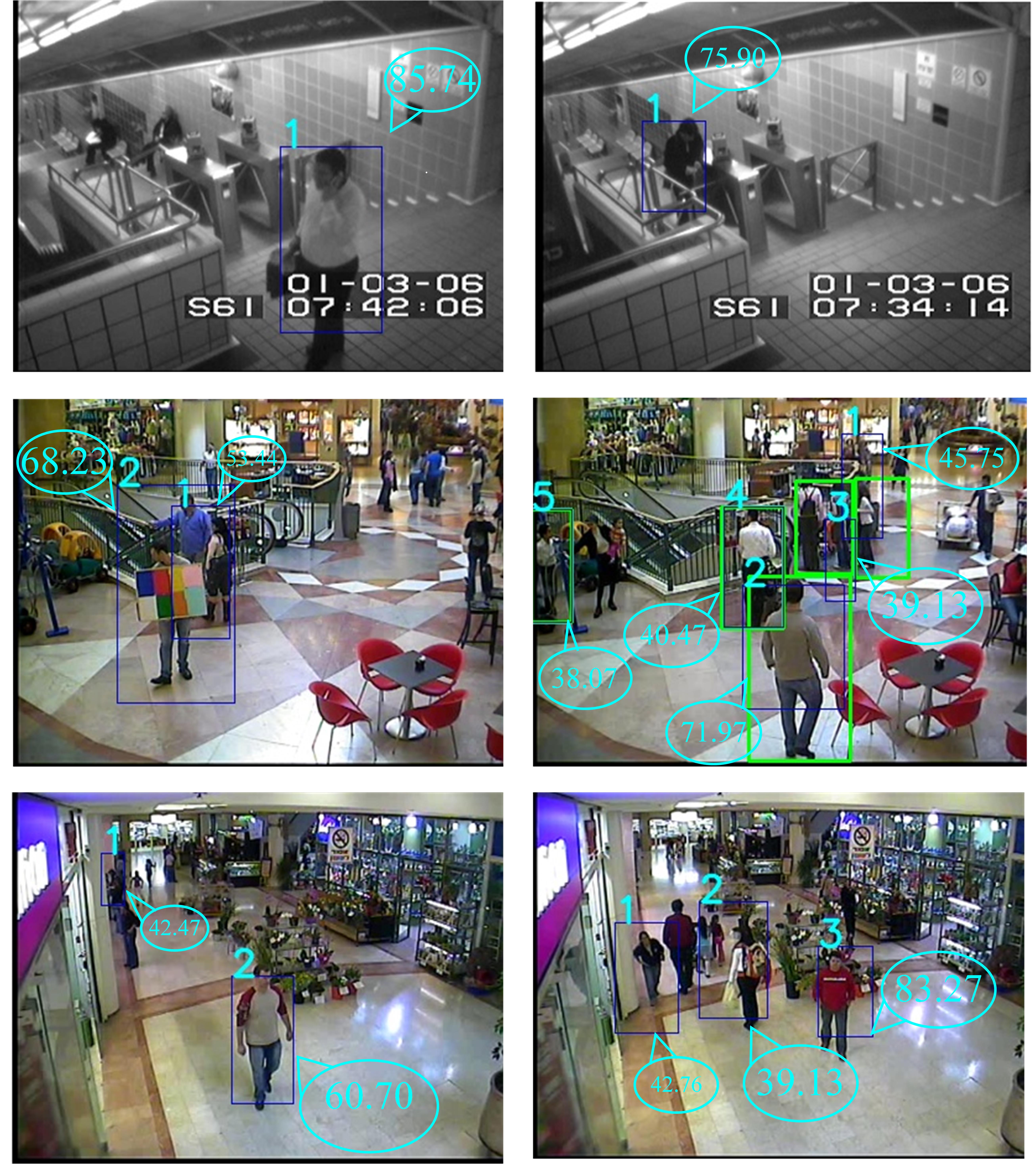}
    \caption{Instances of the fuzzy algorithm output for the videos in the Adam dataset including mall and subway entrance videos.}
    \label{fig:showcase}
    \vspace{-10pt}
\end{figure}

We also ran the I-SAFE on two publicly available video detests, namely Adam \cite{DBLP:journals/pami/AdamRSR08} subway entrance and more than 3 hours of video from a mall security camera. One point which in these datasets is very noticeable, is that the original videos have $\times5$ the number of frames compared to what is provided. The main reason for this down sampling goes back to the slow human motion compared to fast cameras of today. Figure \ref{fig:showcase} shows some of the instances that the I-SAFE detects people and assigns a score to them. In Fig. \ref{fig:showcase}, the bounding boxes around the humans are shown with the corresponding loitering score around each object. Notice how objects that get further away from the camera are not detected, which is the result of inadequate pixel density for detection or in some congested cases low detection resolution. It is worth highlighting that in Fig. \ref{fig:showcase}, the blue boxes are the tracker output while the green boxes show the detection algorithm that checks the frames for new objects with the frequency of 1 in 5 frames.

It can be seen in these samples pictures in Fig. \ref{fig:showcase}, that the camera needs to train to a person with abnormal behavior (here is a man walking on the phone carrying a cardboard piece with different colors on it, for camera detection) for more traditional detection methods.Noted that the proposed I-SAFE does not require training when deployed as camera placement supports the threshold parameters that the operator needs to recalculate. 

Finally, Table \ref{table:result} compares the results of the I-SAFE to the ground-truth and other models for abnormal loitering detection as reported in each paper and shows comparable results. Although the small number of the loitering cases are detected with other algorithms, the I-SAFE achieves these results while minimizing the delay and network overhead in a decentralized edge computing paradigm environment. Observing the scores for the videos in the dataset, we concluded that a threshold of 60\% is best to show the abnormal activity using an average CPU of 68.3\% on a single thread and 96 MB of memory. Of course, with a higher threshold the system gives less False Positives.

Although these examples are used for model performance analysis, they carry very limited number of positive examples to have an otherwise machine-learning solution to them. 

A closer look at Table \ref{table:result} and Fig. \ref{fig:showcase} shows that the human activity recognition heavily depends on how accurately it can detect and track each individual. In the mall video segment, because of the frame complexity and the object partial or complete collision, tracking may be interrupted and data may be lost. The advantage of decoupling feature extraction from decision making is that we can use the same fuzzy model with future more accurate video processing techniques. 

\begin{table}
\begin{center}
\begin{tabular}{|c|c|c|c|c|c|c|}
\hline
\multicolumn{1}{|c|}{\textbf{Detection Model}} & \multicolumn{2}{|c|}{\textbf{Sub. Ent.}} & \multicolumn{2}{|c|}{\textbf{Mall 1}} & \multicolumn{2}{|c|}{\textbf{Mall 2}} \\
\cline{2-7}
& TP&FP & TP&FP & TP&FP \\
\hline\hline
Ground Truth & 14 & 0 & 4 & 0 & 4 & 0 \\
\hline
Adam et al. \cite{DBLP:journals/pami/AdamRSR08} & 13 & \textbf{4} & \textbf{4} & \textbf{1} & \textbf{4} & 3 \\
\hline
Zhao et al. \cite{zhao2011online} & \textbf{14} & 5 & NA & NA & NA & NA \\
\hline
Cocsar et al. \cite{cocsar2017toward} & \textbf{14} & \textbf{4} & NA & NA & NA & NA \\
\hline
Kim et al. \cite{kim2009observe} & 13 & 6 & NA & NA & NA & NA \\
\hline
\textbf{I-SAFE} & 13 & \textbf{4} & 3 & 2 & \textbf{4} & \textbf{1} \\
\hline
\end{tabular}
\end{center}
\caption{Loitering Score in different video samples (TP: True Positive, FP: False Positive, NA: Not Available).}
\label{table:result}
    \vspace{-10pt}
\end{table}

\section{Conclusions}
\label{conc}

The smart surveillance system for public safety should be able to detect suspicious people or activities in realtime. Based on the lightweight human object detection and tracking algorithms previously reported, this paper advances proactive surveillance system design by proposing I-SAFE, an instant suspicious activity identification in the edge paradigm using CNN feature extraction and fuzzy decision making. The algorithms to extract features from incoming video stream are implemented on an edge device, which efficiently reduces the communication overhead and enables outsourcing the decision making process to the fog level. The fog device contextualizes the features, fuses the seven features with a fuzzy logic control system, and provides decision making. The rules and features adopted are chosen under the guidance of campus police officers. A proof-of-concept prototype of the I-SAFE scheme has been implemented and tested using real-world surveillance video streams.

Our on-going efforts consider two directions: (1) Adding features to the tracking and classification algorithms to detect gesture for more accurate decision making; and (2) enhancing the lightweight detection and tracking algorithms to tackle the challenging situations shown in Fig. \ref{fig:fails}.   

\section*{ACKNOWLEDGEMENTS}

The views and conclusions contained herein are those of the authors and should not be interpreted as necessarily representing the official policies or endorsements, either expressed or implied, of the United States Air Force.

\ifCLASSOPTIONcaptionsoff
  \newpage
\fi

\bibliographystyle{IEEEtranS}
\bibliography{L-CNN}

\vfill

\end{document}